\documentclass[runningheads]{llncs}
\usepackage{graphicx}

\usepackage[ruled]{algorithm2e}
\usepackage[colorlinks,
            linkcolor=red,
            anchorcolor=blue,
            citecolor=green]{hyperref}
\usepackage{color}
\usepackage{stfloats}

\begin{document}
\title{Weakly Supervised Segmentation of Vertebral Bodies with Iterative Slice-propagation}

\author{Shiqi Peng\inst{1} 
\and Bolin Lai\inst{1} 
\and Guangyu Yao \inst{2}
\and Xiaoyun Zhang \inst{1}
\and Ya Zhang \inst{1}
\and Yan-Feng Wang \inst{1}
\and Hui Zhao \inst{2}
}

\institute{Cooperative Medianet Innovation Center, Shanghai Jiao Tong University, Shanghai, 200240, PRC \and
Shanghai Jiao Tong University Affiliated Sixth People's Hospital, 200233, PRC \\
\email{pengshiqi@sjtu.edu.cn, lai.b.bryan@gmail.com, ygy504187803@126.com, \{xiaoyun.zhang, ya\_zhang, wangyanfeng, zhao-hui\}@sjtu.edu.cn}}

\titlerunning{Weakly Supervised Segmentation of VBs with Iterative Slice-propagation}
%
%
\authorrunning{S. Peng et al.}
%
\maketitle              
\begin{abstract}

Vertebral body (VB) segmentation is an important preliminary step towards medical visual diagnosis for spinal diseases. However, most previous works require pixel/voxel-wise strong supervisions, which is expensive, tedious and time-consuming for experts to annotate. 
In this paper, we propose a Weakly supervised Iterative Spinal Segmentation (WISS) method leveraging only four corner landmark weak labels on a single sagittal slice to achieve automatic volumetric segmentation from CT images for VBs.
WISS first segments VBs on an annotated sagittal slice in an iterative self-training manner. 
This self-training method alternates between training and refining labels in the training set.
Then WISS proceeds to segment the whole VBs slice by slice with a slice-propagation method to obtain volumetric segmentations.
We evaluate the performance of WISS on a private spinal metastases CT dataset and the public lumbar CT dataset. 
On the first dataset, WISS achieves distinct improvements with regard to two different backbones.
For the second dataset, WISS achieves dice coefficients of $91.7\%$ and $83.7\%$ for mid-sagittal slices and 3D CT volumes, respectively, saving a lot of labeling costs and only sacrificing a little segmentation performance.


\end{abstract}

\vspace{-0.9cm}
\section{Introduction}
\vspace{-0.25cm}

Segmentation of the vertebral bodies (VBs) from CT images is often a prerequisite for many computational spine imaging tasks, such as assessment of spinal deformities, detection of vertebral fractures, and computer-assisted surgical interventions. 
Previous work of VB segmentation can be generally categorized as model based~\cite{vstern2011parametric}, graph theory based~\cite{ali2014vertebral} and machine learning based methods~\cite{Chu_2015}. 
Most recently, convolutional neural networks (CNNs) achieve better performances and thus have been widely used.
U-Net incorporated with VB location prior knowledge for automatic vertebrae segmentation is explored in an iterative fashion~\cite{Lessmann_2018}. 
Nevertheless, all these methods have a same drawback that they require pixel/voxel-wise labeled CT/MR scans to train models. 
It is desired to leverage less fine-grained labels for segmentation so as to save the labeling cost.  

In this paper, we explore a Weakly supervised Iterative Spinal Segmentation (WISS) method for VB segmentation.
Considering that the normal VBs without fractures are roughly quadrilateral from the sagittal view, we thus propose to annotate the position of a VB with four corner landmarks in the mid-sagittal slice and utilize them as weak labels for segmentation.

From any input CT volume with an annotated slice, we first segment the VBs on the annotated slices in a self-training manner~\cite{lee2017deep}.
The segmentation model is improved iteratively by confident prediction selection and densely connected CRF~\cite{densecrf}.
After the segmentation model converges on the mid-sagittal slices, we then adopt a slice-wise propagation method~\cite{cai2018accurate} to generalize this process to other
successive slices to obtain the full volumetric segmentation.
We apply the segmentation model to adjacent slices and create initial masks for them as extra training data, ultimately producing the final segmentations when the model converges. As this process iterates, segmentation results of all CT slices can be obtained and thus 3D VB segmentation is achieved. 


The proposed WISS method for 2D slices is evaluated on a private spinal metastases CT dataset, and the 3D volumetric segmentation is evaluated on the public lumbar spine CT dataset~\cite{Ibragimov_1}.
On the first dataset, WISS achieves an improvement in dice coefficient of $+2.1\%$ and $+3.7\%$ with regard to two different segmentation backbones, respectively. Furthermore, experimental results on random noise disturbed dataset show that WISS is robust to weak and noisy supervision, which is very essential for medical diagnosis.
For the second dataset, WISS achieves dice coefficients of $91.7\%$ and $83.7\%$ on the mid-sagittal slices and 3D CT volumes, respectively.
Compared with state-of-the-art strongly supervised VB segmentation methods~\cite{Lessmann_2018}, WISS saves huge costs of labeling and sacrifices a little segmentation performance, which is very worthwhile in the practical applications.

\vspace{-0.4cm}
\section{Method}
\vspace{-0.25cm}

In the spinal metastases dataset, each CT volume contains a mid-sagittal slice with four corner landmark annotations.
We aim to leverage such weak supervisions on a single slice to achieve the volumetric segmentation.

 
\vspace{-0.3cm}
\subsection{Sagittal Slice Segmentation via Self-training}
\vspace{-0.2cm}

We first connect four corner landmarks to construct the coarse quadrilateral training labels for the mid-sagittal slices.
Then a base segmentation model is trained using these coarse labels.
Such labels are refined by selecting the most confident predictions and recovering the boundaries by fully connected conditional random field (CRF)~\cite{densecrf}, thus additional supervisions are obtained. 
By repeating training and refining procedures, we can achieve better segmentation results and get a better segmentation model.


\vspace{-0.5cm}
\subsubsection*{Segmentation Backbone:}

In our method, Mask R-CNN~\cite{MRCNN} is selected as the segmentation backbone for its quite good performances on object detection task and instance segmentation task. The probability map prediction from Mask R-CNN is utilized for the confidence computation.


Mask R-CNN is a two stage network: the first stage scans image and generates region of interest (ROI) proposals, and the second classifies the proposals and generates bounding boxes and masks. Formally, during training, a multi-task loss is defined on each sampled ROI as
\begin{equation}
\label{eq1}
\mathcal{L} = \mathcal{L}_{cls} + \mathcal{L}_{box} + \mathcal{L}_{mask} + \alpha \mathcal{L}_{edge} .
\end{equation}
The classification loss $\mathcal{L}_{cls}$, bounding-box regression loss $\mathcal{L}_{box}$, and mask loss $\mathcal{L}_{mask}$ are defined in~\cite{MRCNN}. Besides, to preserve the object boundaries, we compared the magnitude and orientation of the edges of the predicted mask with the ground truth. Thus an edge loss~\cite{SLSdeep} is added to the loss function as 
\begin{equation}
\label{eq2}
L_{edge}=\sqrt{(M_x - G_x)^2 + (M_y - G_y)^2} ,
\end{equation}
where $M$ is a generated mask and $G$ is the corresponding ground truth. $(M_x, M_y)$ and $(G_x, G_y)$ are the first derivatives of $M$ and $G$, respectively in $x$ and $y$ directions. 
$\alpha$ is a weighted coefficient. 
Although the edges of the annotations might not be accurate at the beginning, $\mathcal{L}_{edge}$ can be considered as an attention of the possible edge areas. As the iteration proceeds, the annotations will be more and more accurate, thus edge loss will help model converge to the optimal solution.

\vspace{-0.5cm}
\subsubsection*{Confident Prediction Selection:}

\begin{figure}[tbp]
\centerline{\includegraphics[width=\linewidth]{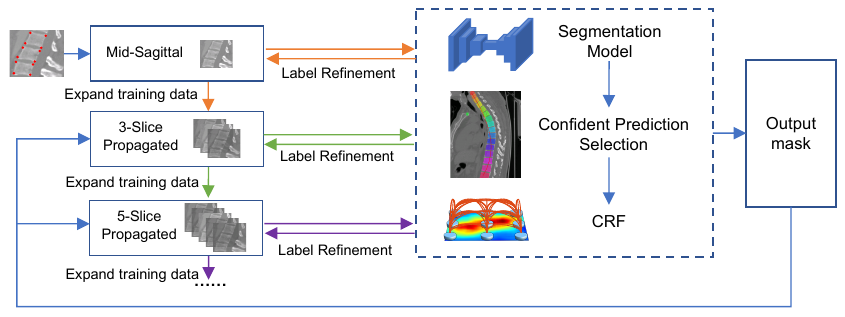}}
\vspace{-0.25cm}
\caption{Overview of the proposed method. We use segmentation output mask incorporating with confident prediction selection and CRF refinements to gradually generate extra training data for VB segmentation. Arrows colored in orange, green and purple represent slice-propagated training at its $1^{st}$, $2^{nd}$ and $3^{rd}$ iterations, respectively. 
For each iteration, training and refining procedures are repeated in a self-training manner until the model converges.
Best viewed in color.} 
\label{fig1}
\end{figure}

We propose a confident prediction selection method to avoid passing errors to the next iteration.
During inference, Mask R-CNN generates three outputs for each predicted ROI: the probability $\mathcal{P}$ to contain a VB, bounding box coordinates $\mathcal{C}$ and a probability map $\mathcal{M}$ of each pixel. 
First, we select the most confident ROIs where the object probability $\mathcal{P}$ exceeds a threshold $\mathcal{T}_1$. 
For each ROI, we select the confident pixels as mask where the probability map $\mathcal{M}$ exceeds another threshold $\mathcal{T}_2$. 
Considering the prior knowledge that VBs are generally arranged in a column, we then fit a curve based on the center points of the confident ROIs.
ROIs close to this curve are selected as final predictions, and ROIs away from this curve are rejected.
As shown in Fig. \ref{fig1}, the colored regions are the predicted ROIs and the red line is the fitted curve. The leftmost green region is a false positive prediction and thus it is discarded.

\vspace{-0.5cm}
\subsubsection*{Error Alleviation by CRF:}

Since the initial coarse quadrilateral labels are imperfect, the output probability maps are not tend to converge to the optimal solution.
We thus adopt the densely connected conditional random fields (CRF) to overcome such a problem.
The CRF model establishes pairwise potentials, which take both pixel positions and intensities into account, on all pairs of pixels in the image. 
Therefore low-level appearances, such as VB boundaries, are incorporated into the refined predictions.
CRF recovers the object boundaries by correcting the oversegmentation regions and undersegmentation regions, thus providing extra training information and alleviating the impact of error amplification.

\vspace{-0.5cm}
\subsection{Slice-Propagated Segmentation}
\vspace{-0.25cm}

In order to generate VB segmentations for all sagittal CT slices, we train the segmentation model in a slice-propagated manner.
Formally, we denote $X_i$, $Y_i$ and $\hat{Y}_i$ as the $i^{th}$ sagittal slices, corresponding labels and the model outputs for all the CT volumes in the training set, respectively. $m$ is the mid-sagittal index.
The segmentation model first learns VB appearances based on the mid-sagittal slices $[X_m]$ via the self-training method introduced above. 
After the model converges, we then apply this model to $[X_{m-1}, X_{m+1}]$ from the entire training set to compute initial predicted probability maps $[\hat{Y}_{m-1}, \hat{Y}_{m+1}]$. 
Given these probability maps, refined VB segmentations $[Y_{m-1}, Y_{m+1}]$ are created using the confident prediction selection and densely connected CRF explained above. 
These segmentations are employed as training labels for the segmentation model on the $[X_{m-1}, X_{m+1}]$ slices, ultimately producing the updated segmentations $[\hat{Y}_{m-1}, \hat{Y}_{m+1}]$ once the model converges. 
As this procedure proceeds iteratively, we can gradually obtain the converged VB segmentation results across all CT slices, and then a volumetric segmentation can be produced by stacking the slice-wise segmentations $[\dots,\hat{Y}_{m-1}, \hat{Y}_{m}, \hat{Y}_{m+1},\dots]$. 
The overview of the proposed method is depicted in Fig. \ref{fig1}.


\vspace{-0.5cm}
\section{Experimental Setup and Results}
\vspace{-0.25cm}

\subsubsection*{Datasets.} 
\textit{The spinal metastases dataset} is used for training and 2D evaluation. All CT scans come from patients with spinal metastases. The scans are reconstructed to in-plane resolution between 0.234mm and 2.0mm, and slice thickness between 0.314mm and 5.0mm. These CT scans cover all of the spine, including cervical, thoracic, lumbar and sacral VBs. 
Training set contains 284 images with 3400 VBs, and testing set contains 95 images with 1109 VBs. The four points annotations and reference segmentations are annotated by three senior radiologists.

\textit{The lumbar spine dataset} consists of 10 scans of healthy subjects of the lumbar vertebraes.  The scans are reconstructed to in-plane resolutions of 0.282mm to 0.791mm and slice thickness of 0.725mm to 1.530mm. We manually edit the reference segmentation to keep only VBs.
The segmentation models are trained on the spinal metastases dataset and directly evaluated on the lumbar dataset.
\vspace{-0.25cm}

\subsubsection*{Evaluation Metrics.} For the spinal metastases dataset, we use Dice coefficient(DIC), Accuracy(ACC), Sensitivity(SEN), and Specificity(SPE) metrics to evaluate the segmentation performances~\cite{ref_dataset2}. Larger values mean better segmentation accuracy. DIC is the most important criterion.

For the lumbar spine dataset, the segmentation performance is evaluated with Dice similarity coefficient (DIC), average symmetric surface distance (ASD), and Hausdorff distance (HSD). 
All metrics are calculated for individual vertebrae and then averaged over all scans. 

\vspace{-0.25cm}
\subsubsection*{Implementation.} The proposed method is implemented using the TensorFlow library \footnote{\href{https://www.tensorflow.org/}{https://www.tensorflow.org/}}. 
Stochastic gradient descent optimizer is used to optimize parameters with a learning rate and momentum of $0.001$ and $0.9$, respectively.
Transfer learning technique is applied since the model is quite large but the training set is relatively small. We adopt weights pre-trained on MSCOCO dataset and only trained the head network, which help the model converge faster and achieve better results. Training for 50 epochs on an NVIDIA TITAN X GPU with 12GB memory takes approximately 2 hours, and testing takes about 0.5s per image. As for self-training, the number of iterations is set to be 2. Confident thresholds $\mathcal{T}_1$ and $\mathcal{T}_2$ are set to be 0.9 and 0.5 respectively. Edge loss coefficient $\alpha$ is set to be 0.1.

\begin{table*}[htbp]
\caption{Performances evaluated on the spinal metastases CT dataset with two segmentation backbones. \textit{M} denotes Mask R-CNN backbone. \textit{E} means using the edge loss (Eq. \ref{eq1}). \textit{R} means using confident prediction selection and fully connected CRF refinements. \textit{ST} stands for self-training. For the last four rows, prefix \textit{n-} represents for experiments on the noisy dataset where the four corner landmark labels are randomly shifted by 0 to 1 millimeter.}
\label{tab1}
\centering
\setlength{\tabcolsep}{3.3mm} {
\begin{tabular}{c|ccccc}
\hline
Methods  & DIC & ACC & SEN & SPE\\
\hline
\hline
UNet\cite{Unet}                  & 86.5          & 95.1          & 79.3          & \textbf{98.8} \\
UNet-ST                          & \textbf{88.6} & \textbf{95.5} & \textbf{86.8} & 97.5 \\
\hline
M\cite{MRCNN}                             & 87.6          & 97.8           & 81.2          & \textbf{99.6} \\
M-E                                       & 87.8          & 97.8           & 82.3          & 99.4 \\
M-E-R                                     & 89.6          & 98.0           & \textbf{89.8} & 98.8 \\
{\bfseries M-E-R-ST}                    & \textbf{90.1} & \textbf{98.1}  & 88.8          & 99.1 \\
\hline
\textit{n-}M                               & 85.6          & 97.4          & 79.6          & 99.3 \\
\textit{n-}M-E                             & 86.2          & 97.6          & 79.4          & \textbf{99.5} \\
\textit{n-}M-E-R                         & 86.2          & 97.6          & 79.5          & \textbf{99.5} \\
{\bfseries \textit{n-}M-E-R-ST}          & \textbf{89.3} & \textbf{97.9} & \textbf{90.6} & 98.6 \\
\hline
\end{tabular} }
\end{table*}

\vspace{-0.25cm}
\subsubsection*{Mid-sagittal Slices Segmentation via self-training.} 
Quantitative results on the spinal metastases CT dataset are shown in Tab. \ref{tab1}. We compare the performances of U-Net~\cite{Unet} and Mask R-CNN~\cite{MRCNN} as segmentation backbones.
As for U-Net backbone, we adopt the same quadrilateral coarse labels, but don't use the confident prediction selection and CRF techniques.
The results demonstrate that our proposed method improves the performances effectively. 
For U-Net backbone, DIC is improved by 2.1\%. 
And for Mask R-CNN backbone, our method outperforms backbone 2.5\% in DIC.  

In practice, errors are inevitable in the manual annotations. 
A generic segmentation method should be robust enough against these labeling noises. 
To prove such robustness of our method, we build a noisy dataset by randomly shifting the four corner landmarks by 0 to 1 millimeter and conduct experiments again on this noisy dataset. 
The experimental results show that though the overall performance is decreased, self-training is still effective. The performance of Mask R-CNN is improved by +2.8\% in DIC. 
The results on this noisy dataset are reported in Tab. \ref{tab1} with prefix \textit{n-}.

\begin{figure}[htbp]
\includegraphics[width=\textwidth]{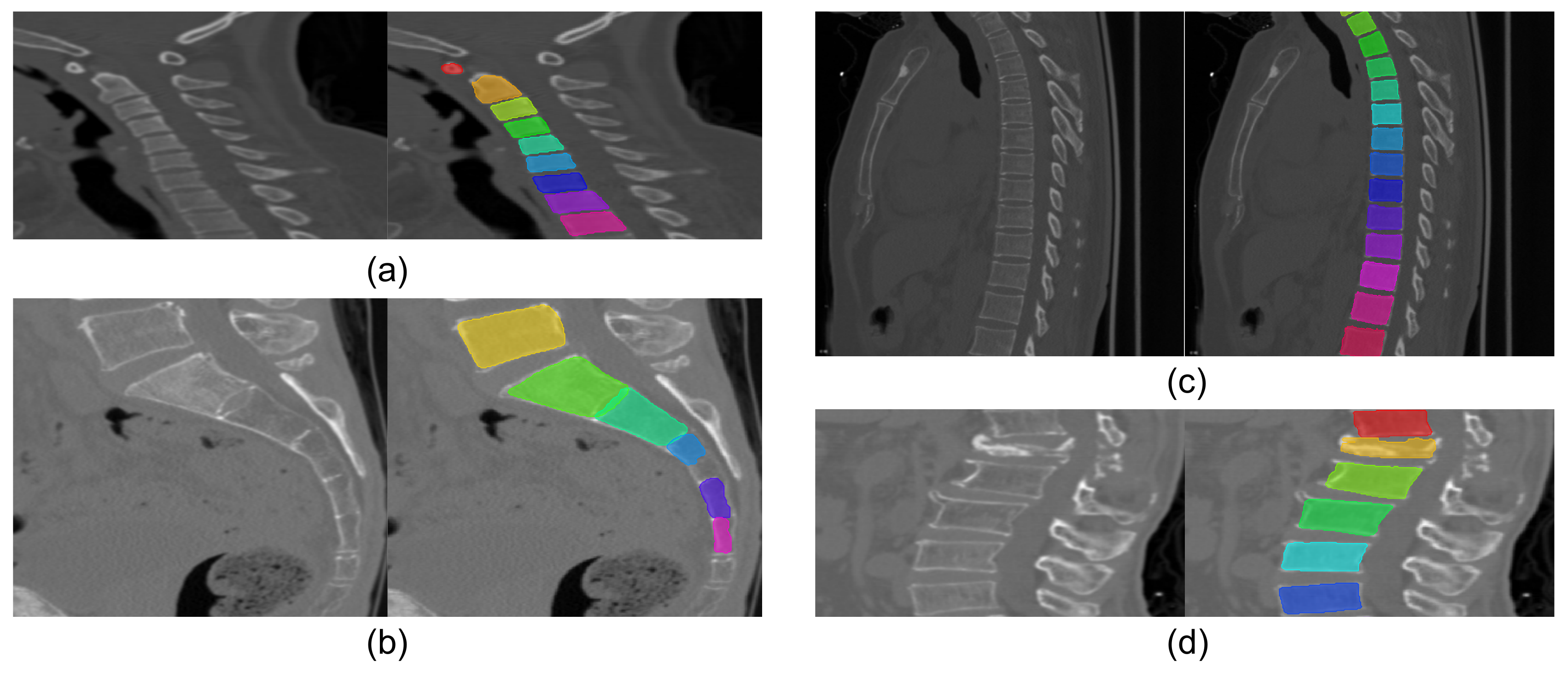}
\vspace{-0.25cm}
\caption{Qualitative results for the spinal metastases dataset. For each case, \textit{left} is the input image and \textit{right} is the segmentation result. \textit{(a) Cervical VBs; (b) Sacral VBs; (c) Thoracic VBs; (d) Lumbar VBs with collapse.} Best viewed in color.} 
\label{fig2}
\end{figure}

Fig. \ref{fig2} shows segmentation results of four typical kinds of images of the spinal metastases dataset. The four examples are cervical, sacral, thoracic and lumbar VBs. Although the appearances of cervical VBs and sacral VBs are quite different from thoracic VBs and lumbar VBs, especially for C1 and C2, our model is robust enough to successfully detect and segment these two difficult cases as demonstrated in Fig. \ref{fig2}.(a). In Fig. \ref{fig2}.(b), two lumbar VBs and five sacral VBs are segmented, but segmentation for S3 is not accurate enough due to its irregular shape. Besides, in Fig. \ref{fig2}.(d), a lumbar VB (the yellow one) suffers metastatic tumor and is therefore collapsed. Its original texture and shape have been destroyed. Even so, our model detects this collapsed VB and segments it successfully, indicating that our model is highly tolerant and robust.

\vspace{-0.5cm}
\subsubsection*{Weakly Supervised Slice-Propagated Segmentation.}

\begin{table*}[htbp]
\caption{Volumetric segmentation results for 10 subjects of the lumbar spine CT dataset.}
\begin{center}
\begin{tabular*}{\hsize}{@{}@{\extracolsep{\fill}}c|cccccccccc|c@{}}
\hline
Metrics &  \#1 & \#2 & \#3 & \#4 & \#5 & \#6 & \#7 & \#8 & \#9 & \#10 & MEAN$\pm$STD\\
\hline
\hline
DIC    & 83.21 & 83.26 & 82.25 & 83.72 & 82.78 & 84.95 & 85.48 & 83.09 & 86.77 & 81.59 & \textbf{83.71$\pm$1.50} \\
\hline
ASD    & 0.50 & 0.50 & 0.51 & 0.48 & 0.49 & 0.43 & 0.52 & 0.50 & 0.46 & 0.88 & \textbf{0.53$\pm$0.12} \\
\hline
HSD    & 4.54 & 5.30 & 4.30 & 3.89 & 4.50 & 4.49 & 5.77 & 4.63 & 4.17 & 4.76 & \textbf{4.64$\pm$0.52} \\
\hline
\end{tabular*}
\label{tab2}
\end{center}
\end{table*}

We conduct the proposed slice-propagated training on the spinal metastases dataset. Then we directly evaluate the 3D segmentation performance on the lumbar spine CT dataset. We report the results with propagating 5 sagittal slices. The dice score for 2D mid-sagittal slices on the lumbar dataset is $91.7\pm2.3$\%, and the volumetric segmentation results are tabulated in Tab. \ref{tab2}. 
We achieve a dice score of $83.71\pm1.50$\%, which is a little lower than the 2D segmentation because the marginal slices is quite difficult than mid-sagittal slices.
State-of-the-art 3D VB segmentation on the lumbar dataset is reported in~\cite{Lessmann_2018} with 96.5\% dice score and 0.2mm ASD. 
It should be noted that our method only bases on four corner landmark weak labels but~\cite{Lessmann_2018} requires strong voxel-wise annotations. 
We save huge labeling costs and achieve volumetric VB segmentation at the cost of a little decrease in performance, which is very worthwhile.

\begin{figure}[htbp]
\includegraphics[width=\textwidth]{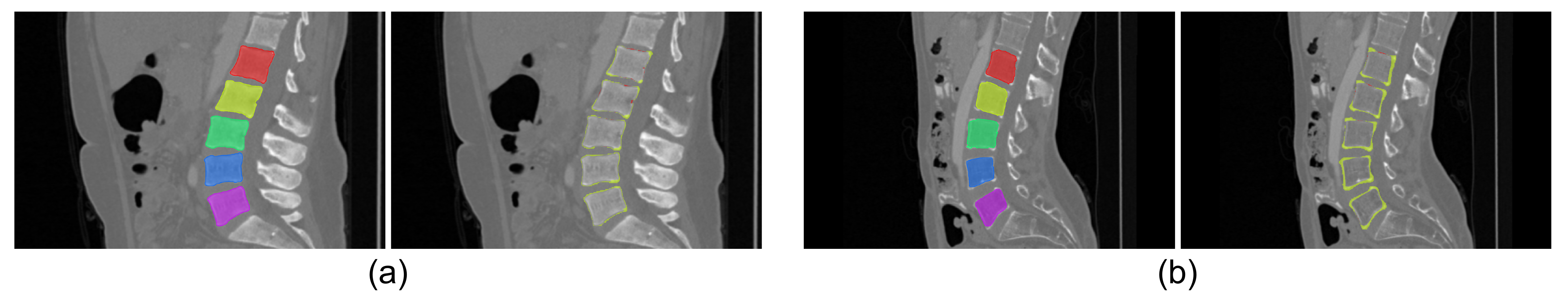}
\vspace{-0.7cm}
\caption{Qualitative results for the lumbar spine CT dataset. (a) \textit{A good case (\#4)}; (b) \textit{A bad case (\#10)}. For each case, left is the segmentation result as color overlay with different colors for different instances, and right is the segmentation results as difference maps with oversegmentation errors marked in red and undersegmentation errors in yellow. Best viewed in color.} 
\label{fig3}
\end{figure}

In addition, we display qualitative segmentation results of mid-sagittal slices for a good case and a bad case in Fig. \ref{fig3}. For the good case, all five lumbar VBs are properly detected and the edges are accurately segmented. However, regarding to the bad case, VBs are successfully detected as well, but some edge regions are missing due to the low image contrast, resulting in a relatively low value of the dice coefficient. 

\vspace{-0.5cm}

\vspace{-0.8cm}
\section{Conclusion}
\vspace{-0.25cm}

In this paper, we proposed a Weakly supervised Iterative Spinal Segmentation (WISS) method leveraging only four corner landmark weak labels on a single sagittal slice to achieve volumetric segmentation from CT images of VBs.
WISS first segments the VBs on the annotated mid-sagittal slices in a self-training manner. Then a slice-wise propagation method is adopted to generalize this process to other successive slices to obtain the full volumetric segmentation.
The experiments have demonstrated that WISS is effective and robust to weak and noisy supervision. 
Furthermore, WISS saves huge labeling costs and only sacrifices a little segmentation performance, which is very valuable in practical applications.
In future works, the proposed WISS method will be applied to other medical applications to prove its versatility.

%
%
\vspace{-0.3cm}

\bibliographystyle{splncs04}
\bibliography{references}

\end{document}